\documentclass[11pt,a4paper]{article}
\usepackage[hyperref]{naaclhlt2019}
\usepackage{times}
\usepackage{latexsym}

\usepackage{url}

\usepackage{amssymb} 
\usepackage{amsmath}
\usepackage{bm}
\usepackage{multirow}
\usepackage{subcaption}
\usepackage{graphicx}  

\newcommand{\vect}[1]{\bm{#1}}
\newcommand{\mat}[1]{\bf{#1}}

\aclfinalcopy 
\def\aclpaperid{*} 

\title{BERT for Joint Intent Classification and Slot Filling}

\author{Qian Chen\thanks{~~Ongoing work.}, Zhu Zhuo, Wen Wang \\
  Speech Lab, DAMO Academy, Alibaba Group \\
  {\tt \{tanqing.cq,~zhuozhu.zz,~w.wang\}@alibaba-inc.com}
}

\date{}

\begin{document}
\maketitle
\begin{abstract}
Intent classification and slot filling are two essential tasks for natural language understanding. They often suffer from small-scale human-labeled training data, resulting in poor generalization capability, especially for rare words. Recently a new language representation model, BERT (Bidirectional Encoder Representations from Transformers), facilitates pre-training deep bidirectional representations on large-scale unlabeled corpora, and has created state-of-the-art models for a wide variety of natural language processing tasks after simple fine-tuning. However, there has not been much effort on exploring BERT for natural language understanding. In this work, we propose a joint intent classification and slot filling model based on BERT. Experimental results demonstrate that our proposed model achieves significant improvement on intent classification accuracy, slot filling F1, and sentence-level semantic frame accuracy on several public benchmark datasets, compared to the attention-based recurrent neural network models and slot-gated models.
\end{abstract}

\section{Introduction}
In recent years, a variety of smart speakers have been deployed and achieved great success, such as Google Home, Amazon Echo, Tmall Genie, which facilitate goal-oriented dialogues and help users to accomplish their tasks through voice interactions. Natural language understanding (NLU) is critical to the performance of goal-oriented spoken dialogue systems. NLU typically includes the intent classification and slot filling tasks, aiming to form a semantic parse for user utterances. Intent classification focuses on predicting the intent of the query, while slot filling extracts semantic concepts. Table~\ref{tab:example} shows an example of intent classification and slot filling for user query ``Find me a movie by Steven Spielberg''.
\begin{table}[ht]
\begin{center}
\scalebox{0.9}{
\begin{tabular}{l| l l}
\hline
\textbf{Query} & \multicolumn{2}{l}{Find me a movie by Steven Spielberg}  \\
\hline
\multirow{3}{*}{\textbf{Frame}} & \textbf{Intent} & find\_movie \\
 & \multirow{2}{*}{\textbf{Slot}} & genre = movie \\
      &  & directed\_by = Steven Spielberg \\
\hline
\end{tabular}
}
\end{center}
\caption{An example from user query to semantic frame.}
\label{tab:example}
\end{table}

Intent classification is a classification problem that predicts the intent label $y^i$
and slot filling is a sequence labeling task that tags the input word sequence $x =(x_1, x_2, \cdots, x_T )$ with the slot label sequence $y^s = (y^s_1 , y^s_2 , \cdots, y^s_T )$. Recurrent neural network (RNN) based approaches, particularly gated recurrent unit (GRU) and long short-term memory (LSTM) models, have achieved state-of-the-art performance for intent classification and slot filling. Recently, several joint learning methods for intent classification and slot filling were proposed to exploit and model the dependencies between the two tasks and improve the performance over independent models~\citep{DBLP:conf/slt/GuoTYZ14,DBLP:conf/interspeech/Hakkani-TurTCCG16,DBLP:conf/interspeech/LiuL16,DBLP:conf/naacl/GooGHHCHC18}. Prior work has shown that attention mechanism~\citep{DBLP:journals/corr/BahdanauCB14} helps RNNs to deal with long-range dependencies.  Hence, attention-based joint learning methods were proposed and achieved the state-of-the-art performance for joint intent classification and slot filling~\citep{DBLP:conf/interspeech/LiuL16,DBLP:conf/naacl/GooGHHCHC18}. 

Lack of human-labeled data for NLU and other natural language processing (NLP) tasks results in poor generalization capability.
To address the data sparsity challenge, a variety of techniques were proposed for training general purpose language representation models using an enormous amount of unannotated text, such as ELMo~\citep{DBLP:conf/naacl/PetersNIGCLZ18} and Generative Pre-trained Transformer (GPT)~\citep{DBLP:techreport/ge1ne8r}. Pre-trained models can be fine-tuned on NLP tasks and have achieved significant improvement over training on task-specific annotated data. More recently, a pre-training technique, Bidirectional Encoder Representations from Transformers (BERT)~\citep{DBLP:journals/corr/abs-1810-04805}, was proposed and has created state-of-the-art models for a wide variety of NLP tasks, including question answering (SQuAD v1.1), natural language inference, and others.

However, there has not been much effort in exploring BERT for NLU. The technical contributions in this work are two folds: 1) we explore the BERT pre-trained model to address the poor generalization capability of NLU; 2) we propose a joint intent classification and slot filling model based on BERT and demonstrate that the proposed model achieves significant improvement on intent classification accuracy, slot filling F1, and sentence-level semantic frame accuracy on several public benchmark datasets, compared to attention-based RNN models and slot-gated models.

\section{Related work}
Deep learning models have been extensively explored in NLU. According to whether intent classification and slot filling are modeled separately or jointly, we categorize NLU models into independent modeling approaches and joint modeling approaches.

Approaches for intent classification include CNN~\citep{DBLP:conf/emnlp/Kim14,DBLP:conf/nips/ZhangZL15}, LSTM~\citep{DBLP:conf/interspeech/RavuriS15}, attention-based CNN~\citep{DBLP:conf/interspeech/ZhaoW16}, hierarchical attention networks~\citep{DBLP:conf/naacl/YangYDHSH16}, adversarial multi-task learning~\citep{DBLP:conf/acl/LiuQH17}, and others.
Approaches for slot filling include CNN~\citep{DBLP:conf/interspeech/Vu16}, deep LSTM~\citep{DBLP:conf/slt/YaoPZYZS14}, RNN-EM~\citep{DBLP:conf/nlpcc/PengYJW15}, encoder-labeler deep LSTM~\citep{DBLP:journals/corr/KurataXZY16}, and joint pointer and attention~\citep{DBLP:conf/acl/ZhaoF18}, among others.

Joint modeling approaches include CNN-CRF~\citep{DBLP:conf/asru/XuS13}, RecNN~\citep{DBLP:conf/slt/GuoTYZ14}, joint RNN-LSTM~\citep{DBLP:conf/interspeech/Hakkani-TurTCCG16},  attention-based BiRNN~\citep{DBLP:conf/interspeech/LiuL16}, and slot-gated attention-based model~\citep{DBLP:conf/naacl/GooGHHCHC18}. 

\section{Proposed Approach}
\begin{figure}[t]
\centering
\includegraphics[width=0.45\textwidth]{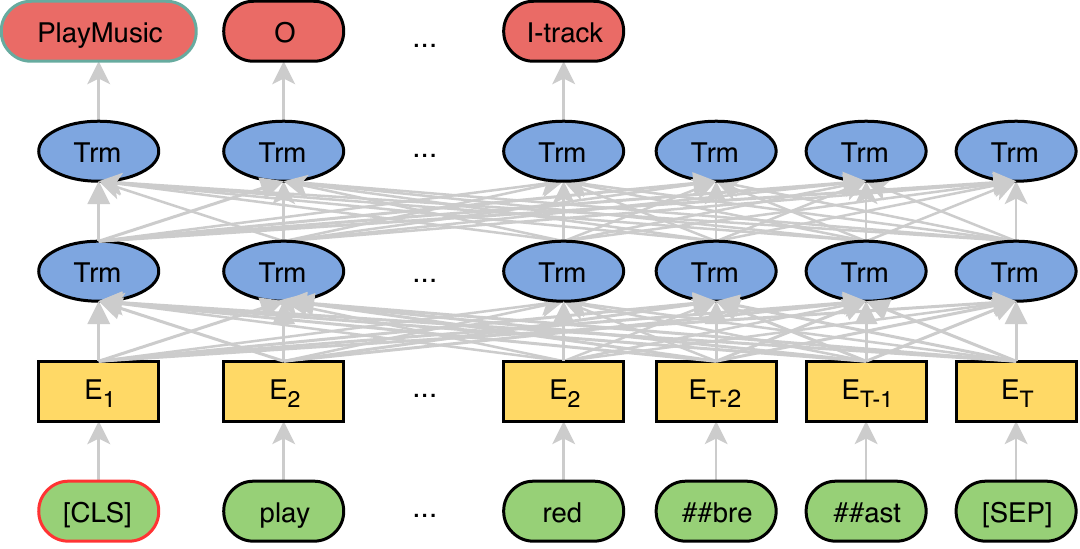}
\caption{A high-level view of the proposed model. The input query is ``play the song little robin redbreast''.}
\label{fig:bert}
\end{figure}

We first briefly describe the BERT model~\citep{DBLP:journals/corr/abs-1810-04805} and then introduce the proposed joint model based on BERT. Figure~\ref{fig:bert} illustrates a high-level view of the proposed model.

\subsection{BERT}

The model architecture of BERT is a multi-layer bidirectional Transformer encoder based on the original Transformer model~\citep{DBLP:conf/nips/VaswaniSPUJGKP17}. 
The input representation is a concatenation of WordPiece embeddings~\citep{DBLP:journals/corr/WuSCLNMKCGMKSJL16}, positional embeddings, and the segment embedding. Specially, for single sentence classification and tagging tasks, the segment embedding has no discrimination. A special classification embedding ([CLS]) is inserted as the first token and a special token ([SEP]) is added as the final token. Given an input token sequence ${\vect x} = (x_1,\dots, x_T)$, the output of BERT is ${\mat H} = ({\vect h}_1,\dots, {\vect h}_T)$.

The BERT model is pre-trained with two strategies on large-scale unlabeled text, i.e., masked language model and next sentence prediction. The pre-trained BERT model provides a powerful context-dependent sentence representation and can be used for various target tasks, i.e., intent classification and slot filling, through the fine-tuning procedure, similar to how it is used for other NLP tasks.

\begin{table*}[ht]
\begin{center}
\begin{tabular}{l c c c c c c}
\hline
\multirow{2}{*}{\textbf{Models}} & \multicolumn{3}{c}{\textbf{Snips}} & \multicolumn{3}{c}{\textbf{ATIS}}  \\
 & \textbf{Intent} & \textbf{Slot} & \textbf{Sent} & \textbf{Intent} & \textbf{Slot} & \textbf{Sent} \\
\hline
RNN-LSTM~\citep{DBLP:conf/interspeech/Hakkani-TurTCCG16} & 96.9 & 87.3 & 73.2 & 92.6 & 94.3 & 80.7 \\
Atten.-BiRNN~\citep{DBLP:conf/interspeech/LiuL16} & 96.7 & 87.8 & 74.1 & 91.1 & 94.2 & 78.9 \\
Slot-Gated~\citep{DBLP:conf/naacl/GooGHHCHC18} & 97.0 & 88.8 & 75.5 & 94.1 & 95.2 & 82.6 \\
\hline
Joint BERT & \textbf{98.6} & \textbf{97.0} & \textbf{92.8} & {97.5} & \textbf{96.1} & {88.2} \\
Joint BERT + CRF & {98.4} & {96.7} & {92.6} & \textbf{97.9} & {96.0} & \textbf{88.6}\\
\hline
\end{tabular}
\end{center}
\caption{NLU performance on Snips and ATIS datasets. The metrics are intent classification accuracy, slot filling F1, and sentence-level semantic frame accuracy (\%). The results for the first group of models are cited from~\citet{DBLP:conf/naacl/GooGHHCHC18}.}
\label{tab:result}
\end{table*}

\subsection{Joint Intent Classification and Slot Filling}
BERT can be easily extended to a joint intent classification and slot filling model. Based on the hidden state of the first special token ([CLS]), denoted ${\vect h}_1$,  the intent is predicted as:
\begin{align}
y^i = \mathrm{softmax}({\mat W}^i {\vect h}_1 + {\vect b}^i) \,,
\end{align}
For slot filling, we feed the final hidden states of other tokens $\vect{h}_2,\dots,\vect{h}_T$ into a softmax layer to classify over the slot filling labels. To make this procedure compatible with the WordPiece tokenization, we feed each tokenized input word into a WordPiece tokenizer and use the hidden state corresponding to the first sub-token as input to the softmax classifier. 
\begin{align}
y^s_n = \mathrm{softmax}({\mat W}^s {\vect h}_n + {\vect b}^s) \,, n \in {1 \dots N}\
\end{align}
\noindent where \({\vect h}_n\) is the hidden state corresponding to the first sub-token of word \(x_n\).

To jointly model intent classification and slot filling, the objective is formulated as:
\begin{align}
p(y^i, y^s|{\vect x}) = p(y^i|{\vect x}) \prod_{n=1}^{N}{p(y^s_n|{\vect x})} \,, 
\end{align}
The learning objective is to maximize the conditional probability $p(y^i, y^s|{\vect x})$. The model is fine-tuned end-to-end via minimizing the cross-entropy loss.

\subsection{Conditional
Random Field }
Slot label predictions are dependent on predictions for surrounding words. It has been shown that structured prediction models can improve the slot filling performance, such as conditional random fields (CRF).~\citet{DBLP:conf/acl/ZhouX15} improves semantic role labeling by adding a CRF layer for a BiLSTM encoder. Here we investigate the efficacy of adding CRF for modeling slot label dependencies, on top of the joint BERT model.

\section{Experiments and Analysis}

We evaluate the proposed model on two public benchmark datasets, ATIS and Snips. 

\subsection{Data}
The ATIS dataset~\citep{DBLP:conf/slt/TurHH10} is widely used in NLU research, which includes audio recordings of people making flight reservations.  We use the same data division as~\citet{DBLP:conf/naacl/GooGHHCHC18} for both datasets.
The training, development and test sets contain 4,478, 500 and 893 utterances, respectively. There are 120 slot labels and 21 intent types for the training set.
We also use Snips~\citep{DBLP:journals/corr/abs-1805-10190}, which is collected from the Snips personal voice assistant. The training, development and test sets contain 13,084, 700 and 700 utterances, respectively. There are 72 slot labels and 7 intent types for the training set.

\subsection{Training Details}
We use English uncased BERT-Base model\footnote{https://github.com/google-research/bert}, which has 12 layers, 768 hidden states, and 12 heads. BERT is pre-trained on BooksCorpus (800M words)~\citep{DBLP:conf/iccv/ZhuKZSUTF15} and English Wikipedia (2,500M words). For fine-tuning, all hyper-parameters are tuned on the development set. The maximum length is 50. The batch size is 128. Adam~\cite{DBLP:journals/corr/KingmaB14} is used for optimization with an initial learning rate of 5e-5. The dropout probability is 0.1. 
The maximum number of epochs is selected from [1, 5, 10, 20, 30, 40].

\subsection{Results}

\begin{table}[ht!]
\begin{center}
\begin{tabular}{l c c c}
\hline
\textbf{Model} & \textbf{Epochs} & \textbf{Intent} & \textbf{Slot} \\
\hline
Joint BERT & 30 & 98.6 & 97.0 \\
No joint & 30 & 98.0 & 95.8 \\
Joint BERT & 40 & 98.3 & 96.4 \\
Joint BERT & 20 & 99.0 & 96.0 \\
Joint BERT & 10 & 98.6 & 96.5 \\
Joint BERT & 5 & 98.0 & 95.1 \\
Joint BERT & 1 & 98.0 & 93.3 \\
\hline
\end{tabular}
\end{center}
\caption{Ablation Analysis for the Snips dataset.}
\label{tab:ablation}
\end{table}

\begin{table*}[ht]
\begin{center}
\scalebox{0.9}{
\begin{tabular}{l p{15cm}}
\hline
\textbf{Query} & need to see \textbf{mother joan of the angels} in one second \\
\hline
\hline
\multicolumn{2}{l}{Gold, predicted by joint BERT correctly} \\
\hline
\textbf{Intent} & SearchScreeningEvent \\
\textbf{Slots} & O O O B-movie-name I-movie-name I-movie-name I-movie-name I-movie-name B-timeRange I-timeRange I-timeRange \\
\hline
\hline
\multicolumn{2}{l}{Predicted by Slot-Gated Model~\cite{DBLP:conf/naacl/GooGHHCHC18}} \\
\hline
\textbf{Intent} & BookRestaurant \\ 
\textbf{Slots} & O O O B-object-name I-object-name I-object-name I-object-name I-object-name B-timeRange I-timeRange I-timeRange \\
\hline
\end{tabular}
}
\end{center}
\caption{A case in the Snips dataset.}
\label{tab:case}
\end{table*}

Table~\ref{tab:result} shows the model performance as slot filling F1, intent classification accuracy, and sentence-level semantic frame accuracy on the Snips and ATIS datasets.

The first group of models are the baselines and it consists of the state-of-the-art joint intent classification and slot filling models: sequence-based joint model using BiLSTM~\citep{DBLP:conf/interspeech/Hakkani-TurTCCG16}, attention-based model~\citep{DBLP:conf/interspeech/LiuL16}, and slot-gated model~\citep{DBLP:conf/naacl/GooGHHCHC18}. 

The second group of models includes the proposed joint BERT models. As can be seen from Table~\ref{tab:result}, joint BERT models significantly outperform the baseline models on both datasets. On Snips, joint BERT achieves intent classification accuracy of 98.6\% (from 97.0\%), slot filling F1 of 97.0\% (from 88.8\%), and sentence-level semantic frame accuracy of 92.8\% (from 75.5\%). On ATIS, joint BERT achieves intent classification accuracy of 97.5\% (from 94.1\%), slot filling F1 of 96.1\% (from 95.2\%), and sentence-level semantic frame accuracy of 88.2\% (from 82.6\%). 
Joint BERT+CRF replaces the softmax classifier with CRF and it performs comparably to BERT, probably due to the self-attention mechanism in Transformer, which may have sufficiently modeled the label structures.

Compared to ATIS, Snips includes multiple domains and has a larger vocabulary. For the more complex Snips dataset, joint BERT achieves a large gain in the sentence-level semantic frame accuracy, from 75.5\% to 92.8\% (22.9\% relative). This demonstrates the strong generalization capability of joint BERT model, considering that it is pre-trained on large-scale text from mismatched domains and genres (books and wikipedia). On ATIS, joint BERT also achieves significant improvement on the sentence-level semantic frame accuracy, from 82.6\% to 88.2\% (6.8\% relative).

\subsection{Ablation Analysis and Case Study}
We conduct ablation analysis on Snips, as shown in Table~\ref{tab:ablation}. Without joint learning, the accuracy of intent classification drops to 98.0\% (from 98.6\%), and the slot filling F1 drops to 95.8\% (from 97.0\%). We also compare the joint BERT model with different fine-tuning epochs. The joint BERT model fine-tuned with only 1 epoch already outperforms the first group of models in Table~\ref{tab:result}.

We further select a case from Snips, as in Table~\ref{tab:case}, showing how joint BERT outperforms the slot-gated model~\cite{DBLP:conf/naacl/GooGHHCHC18} by exploiting the language representation power of BERT to improve the generalization capability. In this case, ``mother joan of the angels'' is wrongly predicted by the slot-gated model as an object name and the intent is also wrong. However, joint BERT correctly predicts the slot labels and intent because ``mother joan of the angels" is a movie entry in Wikipedia. The BERT model was pre-trained partly on Wikipedia and possibly learned this information for this rare phrase.

\section{Conclusion}
We propose a joint intent classification and slot filling model based on BERT, aiming at addressing the poor generalization capability of traditional NLU models. Experimental results show that our proposed joint BERT model outperforms BERT models modeling intent classification and slot filling separately, demonstrating the efficacy of exploiting the relationship between the two tasks. Our proposed joint BERT model achieves significant improvement on intent classification accuracy, slot filling F1, and sentence-level semantic frame accuracy on ATIS and Snips datasets over previous state-of-the-art models. Future work includes evaluations of the proposed approach on other large-scale and more complex NLU datasets, and exploring the efficacy of combining external knowledge with BERT.

\bibliography{naaclhlt2019}
\bibliographystyle{acl_natbib}

\end{document}